\documentclass[10pt,twocolumn,letterpaper]{article}
\usepackage{graphicx}
\usepackage{caption}
\usepackage{subcaption}
\usepackage{cvpr}      

\usepackage{graphicx}
\usepackage{amsmath}
\usepackage{amssymb}
\usepackage{booktabs}

\usepackage[pagebackref,breaklinks,colorlinks]{hyperref}

\usepackage[capitalize]{cleveref}
\crefname{section}{Sec.}{Secs.}
\Crefname{section}{Section}{Sections}
\Crefname{table}{Table}{Tables}
\crefname{table}{Tab.}{Tabs.}

\def\cvprPaperID{30} 
\def\confName{CVPR Workshop NTIRE}
\def\confYear{2023}

\begin{document}

\title{Saliency-aware Stereoscopic Video Retargeting}


\author{Hassan Imani$^{1}$, Md Baharul Islam$^{1,2}$, Lai-Kuan Wong$^{3}$\\
$^{1}$Bahcesehir University \quad \quad \quad
$^{2}$American University of Malta \quad \quad \quad
$^{3}$Multimedia University\\
{\tt\small hassan.imani1987@gmail.com, bislam.eng@gmail.com, lkwong@mmu.edu.my}
}
\maketitle

\begin{abstract}
Stereo video retargeting aims to resize an image to a desired aspect ratio. The quality of retargeted videos can be significantly impacted by the stereo video's spatial, temporal, and disparity coherence, all of which can be impacted by the retargeting process. Due to the lack of a publicly accessible annotated dataset, there is little research on deep learning-based methods for stereo video retargeting. This paper proposes an unsupervised deep learning-based stereo video retargeting network. 
Our model first detects the salient objects and shifts and warps all  objects such that it minimizes the distortion of the salient parts of the stereo frames. We use 1D convolution for shifting the salient objects and design a stereo video Transformer to assist the retargeting process.
To train the network, we use the parallax attention mechanism to fuse the left and right views and feed the retargeted frames to a reconstruction module that reverses the retargeted frames to the input frames. Therefore, the network is trained in an unsupervised manner. Extensive qualitative and quantitative experiments and ablation studies on KITTI stereo 2012 and 2015 datasets demonstrate the efficiency of the proposed method over the existing state-of-the-art methods. The code is available at \url{https://github.com/z65451/SVR/}.

\end{abstract}

\section{Introduction}
\label{sec:intro}
3D video technology is growing in popularity due to the rising demand for augmented and virtual reality (AR/VR) devices  {used in various applications, e.g., mobile phones, autonomous vehicles, and robots.} As 3D videos can be viewed on display devices with varying aspect ratios, stereo image and video retargeting techniques are becoming increasingly important for modifying aspect ratios of media content to correspond to those of target screens and devices. Stereo video retargeting aims to convert a stereo video to the desired aspect ratio. 
Notably, changes in the aspect ratios of videos could result in spatial distortion, 
and temporal inconsistency, such as jittering and flickering. 
Content distortion can be even more severe for stereo videos if depth preservation is not considered during retargeting. Changes in the depth of salient objects can negatively affect 
the 3D viewing experience \cite{li2018depth}. The efficacy of stereo video retargeting approaches depends mainly on the ability to discern between salient and non-salient regions. 

In traditional approaches, the stereo image and video retargeting problem is formulated as a constrained optimization problem.
\cite{bare2015pixel}, \cite{lei2017depth} and \cite{hu2020occlusion} proposed discrete approaches that extends 2D pixel fusion methods for 3D image retargeting. \cite{bare2015pixel}
performs seam-searching by considering depth energy and appearance energy, while in \cite{lei2017depth}, seam selection and seam matching are considered simultaneously
to maintain the relationship between objects and disparity. Hu et al. \cite{hu2020occlusion} combine the occluding masks with the energy optimization of pixel fusion. \cite{li2015depth} introduced a depth-preserving stereo image retargeting technique, a continuous approach. Shao et al. \cite{shao2017qoe} described a Quality of Experience(QoE)-guided warping strategy in response to the effect of QoE on visual attributes. 

\begin{figure*}[htb]
\centering
\includegraphics[width=0.85\linewidth]{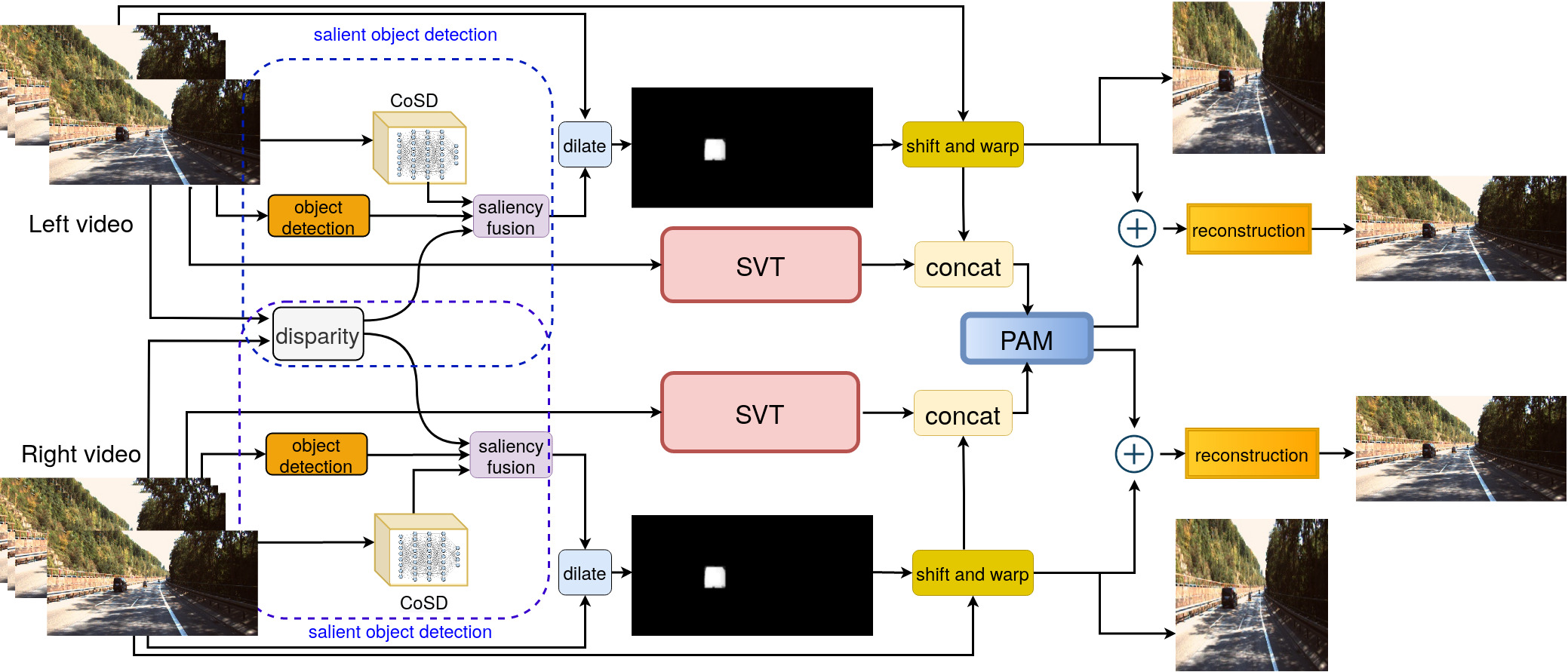}
\vspace{-1em}
\caption{\label{Fig1}
\small
{Proposed stereo video retargeting model architecture. Firstly, using the combination of object detection, disparity information, and {Co-Saliency detection (CoSD), the salient areas of the frames are detected and segmented. The stereo video Transformer {SVT} helps with attention generation. Then, the middle frame is shifted and warped based on the salient regions. The PAM module uses the cross-view information, and a reconstruction block generates the input middle frame.}}}
\vspace{-1em}
\end{figure*}

Kopf et al. \cite{kopf2014warping} proposed one of the first stereo video retargeting methods to preserve salient frame content, avoid flickering, and maintain stereo consistency. 
Liu et al. \cite{liu2015retargeting} formulate distortion energies to prevent significant areas of the videos from deforming. In \cite{islam2019warping}, volume warping with non-homogeneous scaling optimization resizes the stereoscopic video. During the warping, the depth is remapped using a depth remapping constraint and a saliency constraint that protects the salient regions. Temporal and depth constraints are considered in \cite{li2018depth,li2020perceptual}. Li et al. \cite{li2018depth} proposed a method based on depth fidelity constraint. To reduce conflicts between depth, shape, and temporal constraints and prevent perceptually degrading temporal coherence, Li et al. \cite{li2020perceptual} loosen temporal constraints for non-paired regions at frame boundaries. More recently, Wang et al. \cite{wang2021depth} presented a depth trajectory-aware stereoscopic video retargeting technique by optimizing the spatial location and depths, along with a temporal depth distortion energy to preserve the depth trajectory in the temporal direction.

Driven by the proven performance of deep learning in many computer vision tasks, some researchers employed deep neural networks for stereo image retargeting. {\cite{fan2021unsupervised} and \cite{fan2021stereoscopic} proposed convolutional neural network (CNN)-based models to estimate the disparity, {which is then} utilized  to assist in salient objects detection. {Fan et al. \cite{fan2021stereoscopic} created a cross-attention extraction method to build an attention map, and a disparity-assisted 3D importance map preservation module is used to calculate the depth information. 
Fan et al. \cite{fan2021unsupervised} proposed two loss functions for training an unsupervised retargeting model; the view synthesis loss guarantees the generation of high-quality stereoscopic images with inter-view correspondences, and the stereo cycle consistency loss that preserves the structure and prevents disparity variations.} However, the local receptive fields of plain CNN make it difficult to capture correspondence with large disparities \cite{wang2020parallax}. 
To overcome this limitation, Wang et al. \cite{wang2020parallax} integrated epipolar constraints with an attention mechanism to estimate feature similarities along the epipolar line and proposed PAM to handle different stereo frames with extreme disparity changes to cope better with large disparity changes.} To our best knowledge, no research attempted the deep learning approach for stereo video retargeting.

This paper proposes an unsupervised deep learning-based method for stereo video retargeting. Identifying significant stereo video content is essential to retargeting process. In our approach, {we devise a salient object detection scheme that fuses the output of the saliency and object detection models to segment the} important content of the stereo video accurately. To resize the video to the target aspect ratio, we shift and warp the salient content based on the loss of each pixel's shift using a 1D convolutional layer. {We also design the Stereo Video Transformer.} Finally, we re-create the input stereo video frames using cross-view information from the parallax attention mechanism (PAM) \cite{wang2019learning} and propagate the loss to train our model without supervision. Our main contributions are listed as follows:
\vspace{-0.2em}
\begin{itemize}
    \item A novel unsupervised model for stereo video retargeting. By re-creating the input stereo video frames from the retargeted ones, we use the input frames as labels and train the model completely unsupervised.
    \vspace{-0.5em}
    \item A shifting layer that uses convolution and warping for retargeting the video frames.
    \vspace{-0.5em}
    \item A Stereo Video Transformer with self-attention and a sequence of spatial, temporal, and disparity tokens extracted 
    using the stereo patch embedding method. 
    \vspace{-0.5em}
    \item {A loss function that combines the spatial, temporal, and disparity losses to guide the model to obtain a more consistent retargeted stereo video
    .}
\end{itemize}


\section{Proposed Method}
\label{Proposedmethod}
The architecture of the proposed method for stereo video retargeting is shown in Fig. \ref{Fig1}. The input to the framework is a batch of $n$ consecutive left and right frames. 
First, salient objects are detected using disparity information, Co-Saliency detection (CoSD), and object detection techniques. {Then, a dilation operation is applied to expand the salient areas, and the shift-and-warp operation is employed to move the objects in the frames to the appropriate location, given the aspect ratio. To} integrate attention to the model, the proposed stereo video Transformer (SVT) factorizes the input video's spatial, temporal, and depth channels, and its results are concatenated with the warped frames. The PAM module then uses the cross-view information {to fuse both views.} In the reconstruction part, using convolutional blocks, {the objects are relocated to their location within the original aspect ratio, re-generating the input frames,} which is then used to calculate and minimize the loss in the training phase.

\begin{figure}[htb]
\begin{center}
\begin{minipage}{1.0\textwidth}

        \end{minipage}%

  \includegraphics[width=0.12\linewidth]{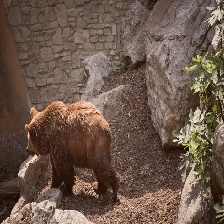}
  \includegraphics[width=0.12\linewidth]{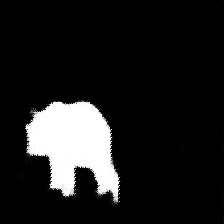}
  \includegraphics[width=0.24\linewidth]{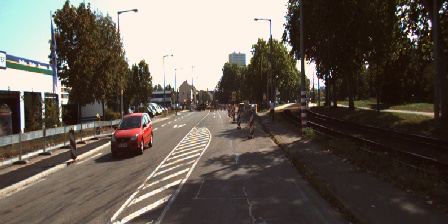}
  \includegraphics[width=0.24\linewidth]{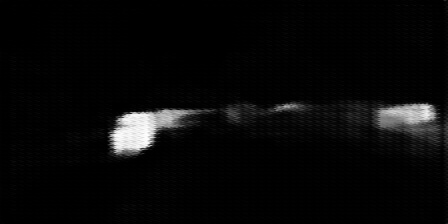}
  \includegraphics[width=0.24\linewidth]{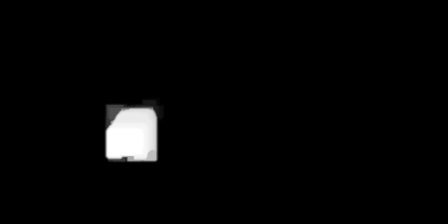}

\end{center}
  \vspace{-1.5em}
  \caption{{Results of saliency detection on the Davis \cite{perazzi2016benchmark} and KITTI stereo 2015 \cite{menze2015joint} datasets. From left to right: image from Davis, its segmentation with CoSD, image from KITTI stereo 2015, its segmentation with CoSD, segmentation with fusion.}}
\label{davis1}
\vspace{-0.5em}
\end{figure}

\subsection{Salient Object Detection}

Detecting salient objects as accurately as possible in a stereo video is an essential step of our model; otherwise, primary object deformation will likely occur. 
Deep neural networks have primarily been trained independently for related tasks such as segmentation and salient object detection, without capitalizing on the inter- and intra-feature cues for a collection of sequential video frames, which may potentially improve the accuracy of object extraction. 

\begin{figure}[htb]
\centering
\includegraphics[width=1.0\linewidth]{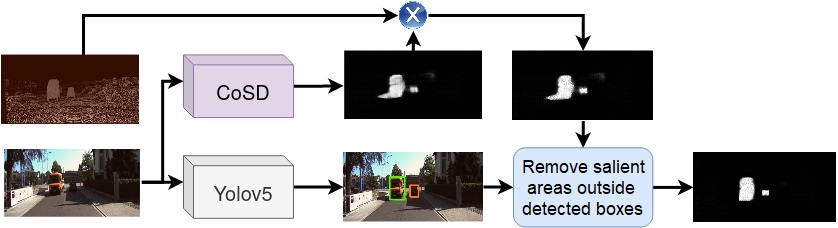}
  \vspace{-1em}
\caption{\label{Fig10}
\small
{Fusion of disparity information, CoSD, and Yolov5 object detection to get better masks for the salient objects.}}
\vspace{-0.5em}
\end{figure}

{Recently, Co-Saliency detection (CoSD), which finds the common salient objects among an image group, is preferred over the normal saliency detection (SD) methods for many computer vision tasks.  CoSD 
discriminate co-occurring objects over consecutive frames \cite{fan2021re} considering other objects in the scene, and both intra-class compactness and inter-class distinctness are maximized simultaneously. } 
Inspired by Su et al.'s \cite{su2023unified} unified framework that jointly detects salient objects and performs segmentation, we adopt their {CoSD} component and combine it with object detection and disparity information to locate the essential areas of a stereo video more accurately. 
The CoSD module contains a transformer block that treats the input frame features as patch tokens and then uses the self-attention technique to extract their long-range dependencies. The network then uses these dependencies to determine the patch-structured similarities between the relevant components. A self-mask is generated using an intra-multi-layer perceptron (MLP) learning module to strengthen the network and prevent partial activation. However, when the camera moves, CoSD alone does not perform well as it fails to detect all salient objects, and some parts of the scene which are not salient are detected as salient areas {(see Figure \ref{davis1})}.
To solve this problem, we propose a fusion strategy that combines CoSD, Yolov5 \cite{ultralytics} object detector, and depth cues from disparity map to generate a more accurate saliency map.
The saliency fusion {block is shown in Figure \ref{Fig10}. We first detect the salient objects with CoSD and fuse its results with the disparity map. Then, we apply Yolov5 to detect the bounding box for each object and remove the salient scene outside of these bounding boxes. This way, we obtain clean salient objects.} 

\subsection{Stereo Video Transformer}

\begin{figure}
\centering
\includegraphics[width=1.0\linewidth]{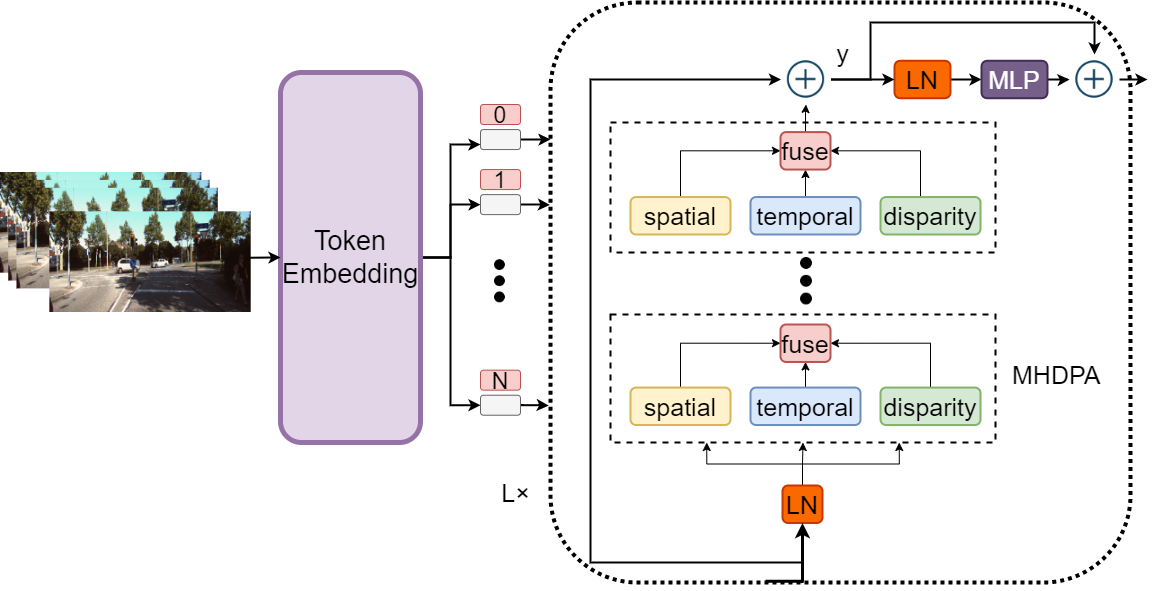}
\caption{\label{FigTrans}
\small 
{Stereo video Transformer} architecture. We factorize all encoder parts into spatial, temporal, and disparity channels.}
\vspace{-1em}
\end{figure}

Attention-based frameworks are rational for modeling long-range contextual relations in the video. Inspired by Vision Transformer (ViT) \cite{dosovitskiy2020image}, which uses a multi-head self-attention mechanism, we propose Stereo Video Transformer (SVT). This model factorizes the stereo video's spatial, temporal, and disparity channels to cope with the long token sequences present in stereo videos. The proposed SVT architecture is shown in Figure \ref{FigTrans}. 

The Transformer has a flexible architecture that works on the provided tokens. ViT 
\cite{dosovitskiy2020image} extracts \textit{N} patches \({f_i}^{h \times w}\) from each video frame \({F}^{H \times W}\) and apply a linear projection to convert them to d-dimensional tokens. 
The tokens are then passed to the Transformer encoder of \textit{L} layers, where each layer \textit{l} contains a multi-head dot-product attention (MHDPA), a layer normalization (LN), and a multi-layer perceptron (MLP).  Let {V=\(\textbf{V}^{T\times H\times W\times C}\)} represent the left or right stereo video. For each patch of size \({t\times h\times w}\) in the \textit{l}th layer, it is mapped to a sequence of tokens, \(\textbf{tokens}_{l}^{n_T\times n_H\times n_W\times d}\), where \({n_t=[T/t]}\), \({n_h=[H/h]}\), and \({n_w=[W/w]}\), and \textit{d} is the token's dimension. 
Instead of sampling \(n_t\) video frames simply as proposed by ViT \cite{dosovitskiy2020image} and concatenating them together independently for the consecutive frames, we propose the stereo patch embedding. {The positional embedding is added to each input token in the same way as the original ViT \cite{dosovitskiy2020image}. The key difference is that the stereo video has more tokens than the pre-trained image
model. Therefore, we initialize the positional embeddings by repeating them temporally.}

\begin{figure}
\centering
\includegraphics[width=0.9\linewidth]{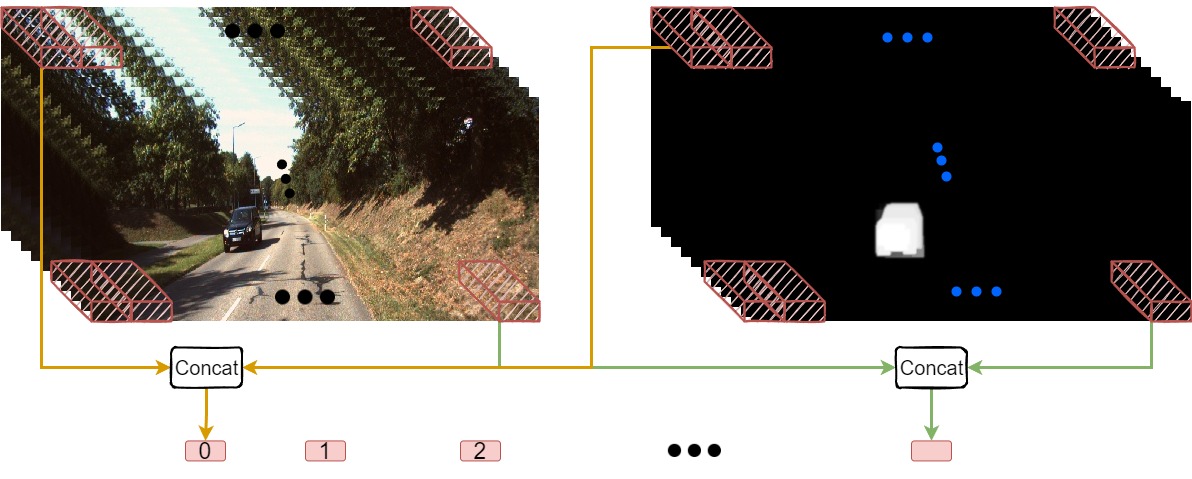}
\caption{\label{FigPatch}
\small 
Patch embedding for stereo video. We extract the spatial, temporal, and disparity patches from each left and right video.}
\vspace{-1em}
\end{figure}

\textbf{Stereo patch embedding.} In addition to spatial and temporal information, depth cues are included for tokenizing the stereo video in the form of disparity information. Figure \ref{FigPatch} illustrates the extraction of spatial, temporal, and disparity patches from the stereo video. For each patch of size \({t\times h\times w}\) with the same disparity size, we first extract 
\({n_t\times n_h\times n_w}\) tokens from the temporal and spatial dimensions. The same process is then performed on the disparity channel. 
Differing from the original ViT \cite{dosovitskiy2020image} where the tokens of the temporal data are combined inside the Transformer encoder, the disparity tokens are fused to the spatial and temporal tokens before feeding them into the Transformer encoder.

\textbf{Spatial, temporal, and disparity self-attention.} We individually calculate attention weights for every token over the spatial, temporal, and disparity channels at various heads. For each head, the attention is as the following:
\begin{equation}
Attn(Q,K,V) = Soft(\dfrac{QK^T}{\sqrt{d_k}} )V
\end{equation}
\noindent where \textit{Attn} and \textit{Soft} refer to the attention and SoftMax, respectively, and the query \(Q=VW_q\), key \(K=VW_k\), and value \(V=VW_v\) are the projection of the left or right video \textit{V}. The primary concept is to create \({(K_s, V_s)^{n_h n_w d}}\), \({(K_d, V_d)^{n_h n_w d}}\), and \({(K_t, V_t)^{n_t d}}\) for spatial, disparity, and temporal indices, respectively, and then adjust the keys and values for each query such that they only look for tokens from the same index. Finally, we concatenate the outputs of different heads with a linear projection.


\subsection{Shift and Warp}

To map the pixels from the left and right source to the left and right target frames, we need to shift the pixels based on the computed {saliency map.} Additionally, we should be aware of the non-salient regions of the frames and properly warp them to avoid the deformation of the non-salient objects. Before shifting, we dilate the salient areas of the frames to recover parts of the salient object that could have been missed.
We first apply a Gaussian blur to the saliency map and then use a 2D convolution with a kernel size of $11\times11$ for dilation. {This size is selected experimentally. We apply this method to each channel.} We then shift the pixels as below:
\begin{equation}
F_{trg}(x,y) = F_{src}(x + shift(x,y), y)
\end{equation} 
\noindent where \textit{\(F_{trg}\)}, \textit{\(F_{src}\), and \textit{shift}} are the retargeted frame, source frame, and the amount of the shift applied to each pixel based on the saliency map of the source frame, respectively. 
The shifting process should constrain the salient regions to be kept as rigid as possible to preserve the salient content.
Additionally, pixels in the same columns should experience comparable shifts to keep the {entire structure of the objects in the frame and prevent deformation on the shape of the main objects.} Therefore, we use a 1D convolution to restrict the shape of a salient area to be consistent along the column axis. The kernel size for this convolution is k=(\textit{\(fr_{height}\)}, 1), where \textit{\(fr_{height}\)} is the height of the frame:
\begin{equation}
S1(x, y) = Conv(F_{src}(x,y), k)
\end{equation}
Next, we calculate the summation of the elements on the y-axis and then tile the elements on the y-axis with dim = (1,\(fr_{height}\),1,1) to get the salient columns:
\begin{equation}
S2(x, y) = Tile(Sum(S1(x,y)), dim)
\end{equation}
\noindent where \textit{Sum} is the summation of elements in the y-axis, and \textit{Tile} constructs a tensor by repeating the elements in the x-axis. The final  shift of each pixel based on the saliency map is calculated as the weighted addition of \textit{S1} and \textit{S2}:
\begin{equation}
shift(x, y) = \alpha S1 + \beta S2
\end{equation}
\noindent where \(\alpha\) and \(\beta\) are experimentally set to $1.9$ and $1$.

Finally, an input frame is warped into the desired aspect ratio using Eq. (2). Four adjacent pixels are interpolated since the shifting map has sub-pixel accuracy.

\subsection{Parallax Attention Mechanism}
\label{Modified PAM}
Based on self-attention approaches \cite{zhang2019self,fu2019dual}, Wang et al. \cite{wang2019learning} introduced the parallax attention mechanism (PAM) to determine matching in stereo images. PAM effectively merges the characteristics of the left and right image pair. The PAM structure has been modified in \cite{imani2022new} to make it suitable for video-based inputs. {After applying a $1\times1$ convolutional layer, the attention mappings from the left to right and vice versa are} built using batch-wised matrix multiplication in a SoftMax block. Features for the left disparity are then merged with the corresponding right features at all disparity levels. To generate more features and increase learning capacity, we use the method in \cite{imani2022new}, in which 2D CNNs are applied to the output features, followed by a ReLU and a batch normalization (BN) layer. Three convolution layers are used with $128$, $128$, and $64$ filters.

\subsection{Reconstruction}

After applying the PAM module, we re-generate the input stereo video frames. For this purpose, we use $5$ 2D CNN blocks, with $64$, $128$, $512$, $128$, and $3$ output filters, respectively. We use the last layer of this block for loss calculation. {The first layer accepts the addition of the outputs of PAM and shifted and warp modules and produces a feature map} of size $64$. Its kernel size is $5$. The other convolution layers use a kernel size of $3$. A stride and padding of $1$ and  ReLU function are used for each  convolutional layer.






\subsection{Loss Functions}

Pixel-based metrics such as L2 or logistic regression are often utilized to determine the loss between the source and recreated frames. 
However, pixel-based loss functions may not accurately represent the subjective difference and spatial relationship between two consecutive frames. For instance, a similar frame that has been moved a few pixels may not substantially impact human perception, but its pixel-by-pixel loss can be severe.

We combine four loss functions for training the proposed model on the KITTI stereo 2012 \cite{geiger2012we} and 2015 \cite{menze2015joint} datasets. The first loss computes the difference between the source and the retargeted frames. The second loss computes the dissimilarities between the source and output of the reconstruction module. The third loss measures the difference between the disparity of the source and retargeted frames. Please note that the aspect ratio of the source and retargeted frames are different. For example, for $50\%$ resizing, the source frames are with size $224\times448$, and the retargeted frames are $224\times224$. 

The first loss computes the difference between the VGG19 \cite{simonyan2014very} features extracted from the source and retargeted frames. Specifically, we use VGG19 features of layers conv1\_2, conv2\_2, conv3\_3 feature before the ReLU activation layer:
\begin{equation}\
L_{VGG19} = MSE(VGG19_{src} - VGG19_{ret})
\end{equation}
\noindent where \textit{MSE} denote the mean square error. 
The total VGG19 features loss is computed as the summation of the feature difference (1) between the entire frames of source and retargeted frames, 
and (2) between their salient regions: 
\begin{equation}\
L_{VGG19}^{total} = L_{VGG19}^{entire} + L_{VGG19}^{salient}
\end{equation}

The second loss term computes the frequency domain differences by computing the MSE between the forward and inverse 2D discrete wavelet transform (DWT) decompositions between source and retargeted frames:
\begin{equation}\
\begin{split}
L_{DWT} = MSE(FDWT_{src} - FDWT_{ret}) \\
+ MSE(IDWT_{src} - IDWT_{ret})
\end{split}
\end{equation}
\noindent where \textit{FDWT} and \textit{IDWT} denote the forward and inverse 2D DWT decompositions, respectively. {\(L_{DWT}\)} is calculated as the average loss of the left and right frames.


The final loss functions are the photometric \(L_{p}\) and smoothness \(L_{s}\) losses \cite{wang2020parallax} respectively. The photometric loss includes a mean absolute error (MAE) loss and a structural similarity index (SSIM) loss term. {The photometric loss is defined as follows:}
\begin{equation}
\begin{split}
L_{p} =\frac{1} N  \sum_{p\in V_{l}}^{}\mathop{}_{\mkern-5mu } \gamma \frac{1-S(I_l (p), \hat{I}(p))} 2 + (1-\gamma) ||I_l (p), \hat{I}(p)||
\end{split}
\end{equation}
\noindent {where \(\hat{I}\) is the warped version of the right frame. S is the SSIM operator, p indicates a valid pixel covered by the valid mask, N is the number of valid pixels, and \(\gamma\) is a constant.}

The smoothness loss is an edge-aware loss that encourages local smoothness of the disparity values:
\begin{equation}
\begin{split}
L_{s} =\frac{1} N  \sum_{p}^{}\mathop{}_{\mkern-5mu } (|| \nabla _x \hat{D}_{r}(p) || e^{-|| \nabla _x \hat{I}_{l}(p) ||} + \\
|| \nabla _y \hat{D}_{r}(p) || e^{-|| \nabla _y \hat{I}_{l}(p) ||}) 
\end{split}
\end{equation}
\noindent {where \(\nabla\) is the gradient operator. The final loss function is the union of the losses above:}
\begin{equation}
\begin{split}
loss = L_{VGG19}^{total} + \alpha L_{DWT}+L_{s} + L_{p}
\end{split}
\end{equation}
\noindent where \(\alpha\) is the regularization term empirically set {to} $0.05$.

\section{Datasets and Experiments}
\label{exp}


\textbf{Datasets: } {Due to the nature of the SVR, no publicly available dataset is specifically designed for the SVR task. Some works, such as cite{li2020perceptual}, used videos from commercial 3D movie films for experiments. The difference between our method and \cite{li2020perceptual} (e.g., not a learning-based method) is the training phase that we have. 
Based on the available existing stereo video datasets, we chose the KITTI stereo 2012 \cite{geiger2012we} and 2015 \cite{menze2015joint} datasets because of the large disparity range between foreground and background objects with significant temporal disparity changes. The video scenes are dynamic, and the camera is moving. However,
our method can also retarget the commercial stereo video (single-scene) to the target aspect ratio.} 
The KITTI stereo 2012 dataset contains $194$ training frame pairs and $195$ test frame pairs. 
There are $200$ training and $200$ test sequences in the KITTI stereo 2015 benchmark (4 frames per scene). 
Since the disparity values are published only for the training sets, we divide their training sets into the train-test sets with an 80:20 split ratio 
and use them for the experiments. 

\textbf{Experiments: }
The specification of the computer system used for our experiments is Intel i7-10875H, 64GB memory, Nvidia RTX3090 24GB. 
We trained our model with ADAM optimizer, 
learning rate initialized to $0.05$ and the model waas trained with $4000$ iterations. {The training took $2$ days to complete on our RTX3090 GPU.}

\textbf{Evaluation Criteria: }
We use qualitative and quantitative comparisons to evaluate the proposed method. For qualitative and quantitative studies, we compare our method with $4$ other methods: linear scaling, manual cropping, fast video \cite{chuning2019fast}, and seam carving \cite{avidan2007seam} methods. We cannot compare our results with the stereo video retargeting techniques \cite{kopf2014warping,liu2015retargeting,li2018depth,islam2019warping} due to the unavailability of codes, including two more recent 2D video retargeting methods  \cite{lee2020object,tan2019cycle}.  
For quantitative comparisons, we use three metrics. The first one is the bidirectional similarity metric \cite{simakov2008summarizing}, a frequently used metric in the image and video retargeting. We compare different retargeting methods for the second metric based on the perceptual distance between the source and retargeted stereo video's VGG19 \cite{simonyan2014very} features. In \cite{li2018depth}, an objective metric named Disparity Distortion ratio (DDr) is proposed to quantify the spatial and temporal depth distortion. We use DDr to compute the mean of the disparity variation of pixels between the retargeted and original videos, normalized to the disparity range as follows:
\begin{equation}\
DDr = \frac{1}{|d_{max}|\times H\times W \times T} \sum{(D - \tilde{D})}
\end{equation}
\noindent where \(d_{max}\) is the maximum disparity in the source stereo video, \textit{H}, \textit{W}, and \textit{T} are the dimensions of the video, and \textit{D} and \(\tilde{D}\) is the disparity maps of the source and retargeted videos, respectively.

\begin{figure*}[htb]
\begin{center}
\begin{minipage}{1.0\textwidth}

        \end{minipage}%
        

  \includegraphics[width=0.26\linewidth]{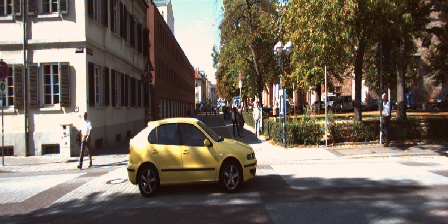}
  \includegraphics[width=0.13\linewidth]{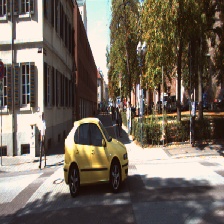}
  \includegraphics[width=0.13\linewidth]{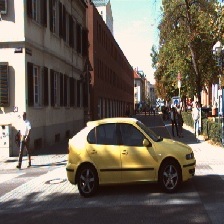}
  \includegraphics[width=0.13\linewidth]{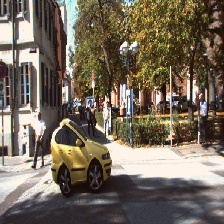}
  \includegraphics[width=0.13\linewidth]{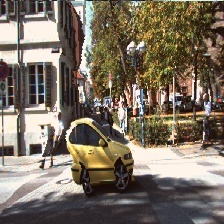}
  \includegraphics[width=0.13\linewidth]{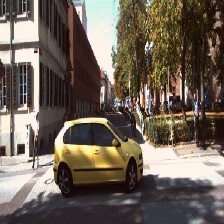}
  \vspace*{0.1cm}

  \includegraphics[width=0.26\linewidth]{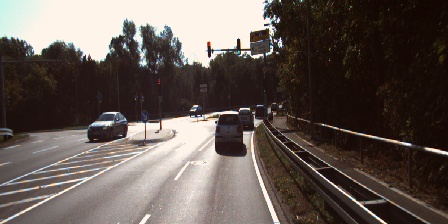}
  \includegraphics[width=0.13\linewidth]{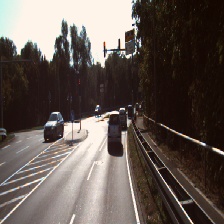}
  \includegraphics[width=0.13\linewidth]{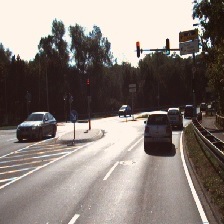}
  \includegraphics[width=0.13\linewidth]{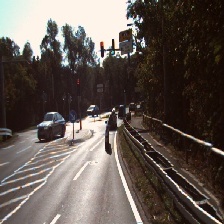}
  \includegraphics[width=0.13\linewidth]{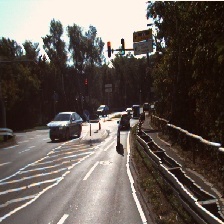}
  \includegraphics[width=0.13\linewidth]{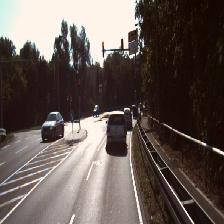}
  \vspace*{0.05cm} 

 \includegraphics[width=0.26\linewidth]{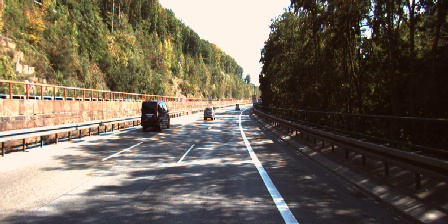}
  \includegraphics[width=0.13\linewidth]{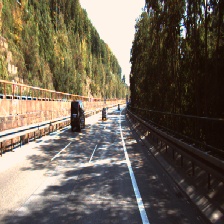}
  \includegraphics[width=0.13\linewidth]{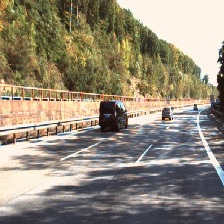}
  \includegraphics[width=0.13\linewidth]{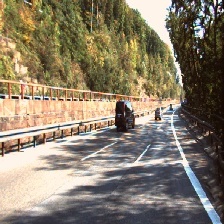}
  \includegraphics[width=0.13\linewidth]{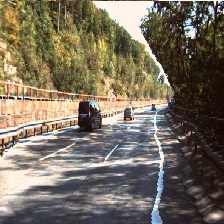}
  \includegraphics[width=0.13\linewidth]{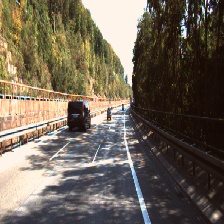}

  \vspace*{0.05cm} 

 \includegraphics[width=0.26\linewidth]{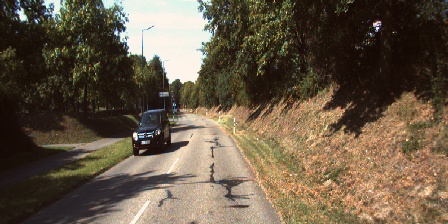}
  \includegraphics[width=0.13\linewidth]{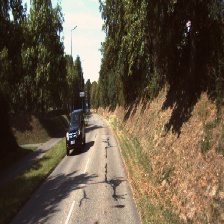}
  \includegraphics[width=0.13\linewidth]{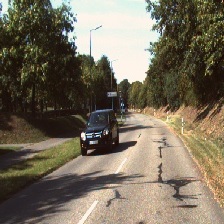}
  \includegraphics[width=0.13\linewidth]{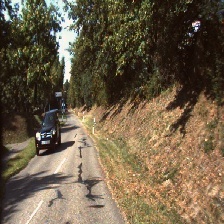}
  \includegraphics[width=0.13\linewidth]{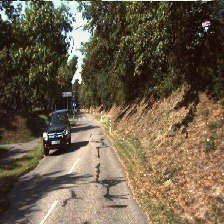}
  \includegraphics[width=0.13\linewidth]{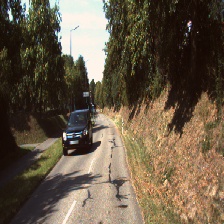}


   
\end{center}
\vspace{-1.5em}
  \caption{{Qualitative results of stereo video retargeting on randomly selected left video frames from the KITTI stereo 2015 \cite{menze2015joint} (top) and 2012 \cite{geiger2012we} (bottom) datasets for reducing the horizontal video size at 50\%. Left to right: original frame, linear scaling, manual cropping, seam carve \cite{avidan2007seam},  fast video \cite{chuning2019fast}, and ours.}}
\label{d1}
\vspace{-1em}
\end{figure*}

\begin{figure}[b] 
\begin{center}

        
  \includegraphics[width=0.245\linewidth]{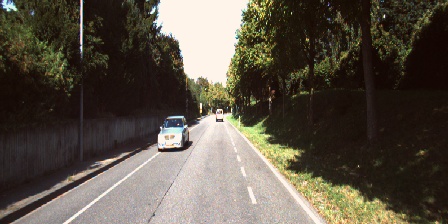}
  \includegraphics[width=0.175\linewidth]{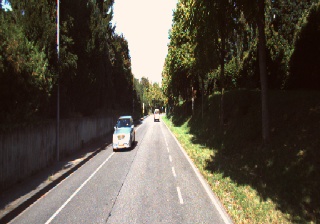}
  \includegraphics[width=0.175\linewidth]{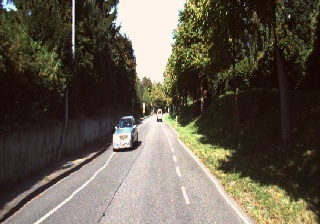}
  \includegraphics[width=0.175\linewidth]{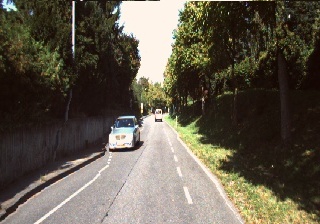}
  \includegraphics[width=0.175\linewidth]{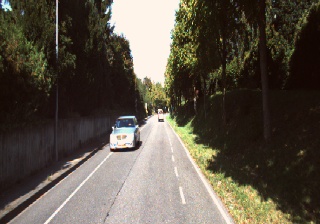}
  \vspace*{0.1cm}




  \includegraphics[width=0.22\linewidth]{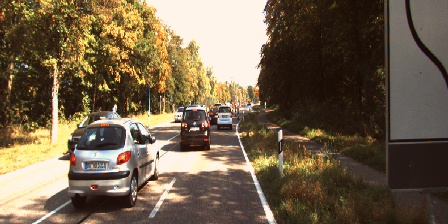}
  \includegraphics[width=0.185\linewidth]{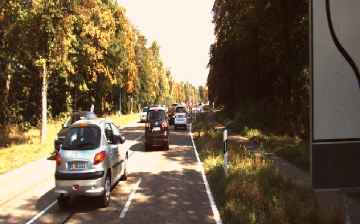}
  \includegraphics[width=0.185\linewidth]{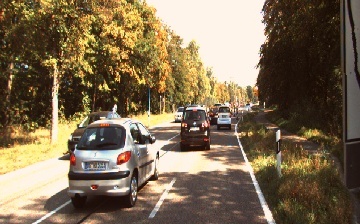}
  \includegraphics[width=0.185\linewidth]{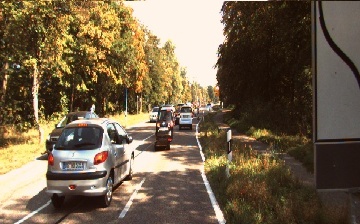}
  \includegraphics[width=0.185\linewidth]{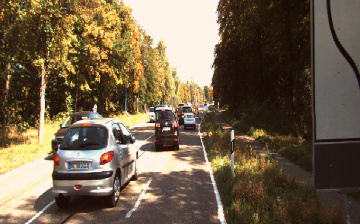}

   \vspace*{-0.5cm} 
\end{center}
  \caption{Qualitative results of retargeting on randomly selected frames from the KITTI stereo 2015 \cite{menze2015joint} dataset for 30\% (first row) and 20\% (second row) reduced the horizontal size. From left to right: input frame, LS, seam carve \cite{avidan2007seam}, fast video\cite{chuning2019fast}, and Ours.}
\label{d3}
\vspace{-1em}
\end{figure}




  

\begin{figure}[b]
\begin{center}
\begin{minipage}{1.0\textwidth}

\hspace{0.4cm} $DDr=$ \hspace{0.5cm} $0.226$  \hspace{0.5cm} $0.112$  \hspace{1.3 cm} $0.233$
        \end{minipage}%

  \includegraphics[width=0.260\linewidth]{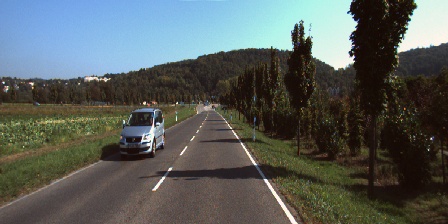}
  \includegraphics[width=0.132\linewidth]{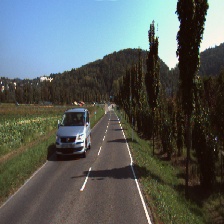}
  \includegraphics[width=0.205\linewidth]{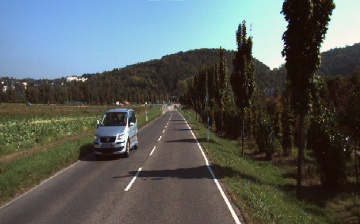}
  \includegraphics[width=0.368\linewidth]{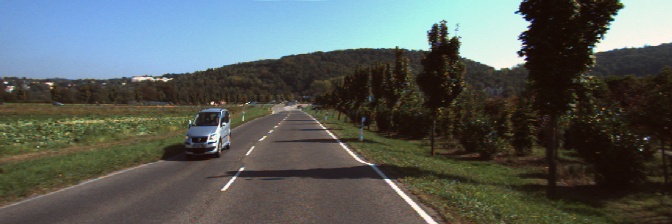}
  \vspace{-0.05cm}
  \includegraphics[width=0.260\linewidth]{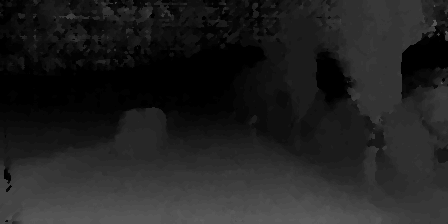}
  \includegraphics[width=0.132\linewidth]{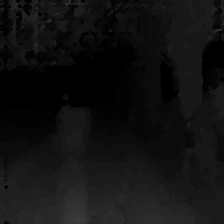}
  \includegraphics[width=0.205\linewidth]{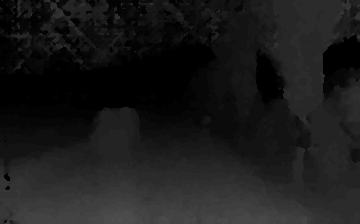}
  \includegraphics[width=0.368\linewidth]{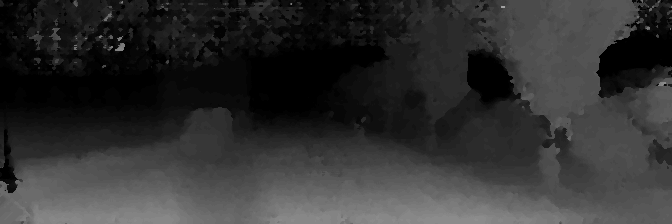}
  
\end{center}
  \vspace*{-0.5cm} 
  \caption{Different retargeting results with respective depth maps. Left to right: Input video frame and their retargeted results with horizontal size reduction at $50\%$, {$20\%$,} and $150\% (enlarge)$.}
\label{d4}
\vspace{-1em}
\end{figure}




\section{Results and Discussion}
\label{Results and Discussions}

\subsection{{Computational Performance}}
{Table \ref{table3} compares the computational complexity of our method to that of other methods for a single pair of frames by computing the sum of the running times for the left and right frames. All methods are implemented on an Intel i7-10875H workstation with Nvidia RTX3090 GPU. We can observe that our proposed method achieved the fastest speed, about 30x and 2x times faster than Seam carving \cite{avidan2007seam} and Fast video \cite{chuning2019fast} methods, respectively.} 

\begin{table}[h!]
		\centering
			\caption{{Comparison of computational complexity. 
            }}
            \vspace{-0.8em}
		\scalebox{0.85}{
		\begin{tabular}{l@{\hspace{6pt}} *{13}{c}}
			\hline 
            {{method}} & Seam carving \cite{avidan2007seam} & Fast video \cite{chuning2019fast} 
            & Ours (GPU)  \\
			\hline
			complexity(s)   & 42.184 &  3.014 
            & 1.831
			\\

			\hline 
		\end{tabular}}
	\label{table3}
\end{table}

\subsection{Qualitative Results}

We randomly select stereo video sequences from KITTI stereo 2012 \cite{geiger2012we} and 2015 \cite{menze2015joint} test sets for our qualitative comparisons. We provide the qualitative results for $3$ aspect ratios: reduction of horizontal video size by {$20\%$, $30\%$,} and $50\%$ respectively. For example, the aspect ratio of $20\%$ means that the size of the original video frames is reduced by 20\% horizontally. Figure \ref{d1} compares the proposed method's results with $4$ methods for $50\%$ aspect ratio on $3$ randomly selected videos from the KITTI stereo 2015 \cite{menze2015joint} test set. Each row belongs to one left frame of one of the videos in the dataset. Since both the left and right views need more space, we report the results based on the left frames and provide all results in the supplementary materials. Each stereo video contains one main object (foreground) and the background. Since KITTI stereo 2012 \cite{geiger2012we} and 2015 \cite{menze2015joint} datasets mostly contain the car videos, the foreground object of the stereo videos includes the cars. It is apparent from visual results in this figure that our method can preserve both the salient object and the background well. Noticeably, the main object size is resized less than the background. 

Figure \ref{d3} shows the retargeting results with horizontal size reduction of $30\%$ and $20\%$. These visual results demonstrate that our method is superior to the other methods. These two aspect ratios require a lesser {shift of pixels. Thus,} the main structure of the video frames is well-preserved by all methods. When we resize the frames with a higher reduction in size, e.g., $50\%$, shape deformation of objects is observed in some methods. Figure \ref{d4} illustrates the results of retargeting one video from KITTI stereo 2012 \cite{geiger2012we} dataset with horizontal video size reduction of $50\%$, {$20\%$,} and $150\%$ (enlarge). It is apparent in these results that our method can effectively perform stereo retargeting in all cases, from very extreme ($50\%$ and $150\%$) to lower resizing ratio {($20\%$)} cases. In addition, this figure's results demonstrate our method's ability to enlarge the frames. 

\begin{table}[b] 
		\centering
			\caption{Comparison of the bidirectional similarity metric \cite{simakov2008summarizing}. 
            The results are the average values for the left and right frames. Videos \#1 and \#2 are taken from KITTI 2015 \cite{menze2015joint}, and videos \#3 and \#4 from 2012 \cite{geiger2012we} datasets. The last column shows the average bidirectional similarity. 
            }
            \vspace{-0.5em}
		\scalebox{0.87}{
		\begin{tabular}{l@{\hspace{25pt}} *{15}{c}}
			\hline 
            {\textbf{Video No.}} &  {\textbf{\#1}} & {\textbf{\#2}} & {\textbf{\#3}} & {\textbf{\#4}}&  {\textbf{Avg}} \\
			\hline
			Manual cropping  & 5.065 & 4.950   & 2.170  & 4.320 & 4.126 \\
			Seam carving \cite{avidan2007seam}  & 3.347 & 2.693  & 3.822  & 3.011 & 3.218 \\
			Fast video \cite{chuning2019fast}  & 3.211 & 2.901   & 3.736  & 2.882 & 3.182\\
			\textbf{Ours}  & \textbf{2.190} & \textbf{1.970}   & \textbf{2.588}  & \textbf{1.870} &\textbf{ 2.154}\\

			\hline 
		\end{tabular}}
	\label{table1}
 \vspace{-0.5 em}
\end{table}

\begin{table}[htb]
		\centering
			\caption{Comparison of the similarity between the input and retargeted videos based on VGG19 \cite{simonyan2014very} feature difference. 
            }
            \vspace{-0.5em}
		\scalebox{0.87}{
		\begin{tabular}{l@{\hspace{25pt}} *{15}{c}}
			\hline 
            {\textbf{Video No.}} & {\textbf{\#1}} & {\textbf{\#2}} & {\textbf{\#3}}& {\textbf{\#4}} & {\textbf{Avg}}\\
			\hline
			Manual cropping  & 1.140 & 1.374 & 1.016 & 0.888  & 1.104\\
			Seam carving \cite{avidan2007seam}  & 0.595 & 1.137 & 0.796 & 0.776 & 0.826\\
			Fast video \cite{chuning2019fast}  & 0.592 & 0.940 & 0.743 & 0.690 & 0.741\\
			\textbf{Ours}  & \textbf{0.326} & \textbf{0.739} & \textbf{0.721} & \textbf{0.678} & \textbf{0.616}\\
			\hline 
		\end{tabular}}
	\label{table2}
\end{table}


\subsection{Quantitative Results}
We use two methods for computing the similarity between two frames in our quantitative comparisons. The first method is the bidirectional similarity metric \cite{simakov2008summarizing}, the most widely accepted criteria for assessing the video retargeting quantitative performance \cite{cho2020temporal}. When evaluating the quality of the retargeted video, bidirectional similarity looks at the coherence and completeness between the source and retargeted frames. Completeness assesses the impairment in the shape of the retargeted objects relative to the objects in the source frames. In contrast, coherence measures the deformity that occurs when an area that does not exist in the original frames appears in the retargeted frames.

Table \ref{table1} compares the bidirectional similarity between the source and target videos. A total of $4$ videos are used for this study. A lower value represents better shape preservation or less object deformation during the retargeting. Our method achieves the best results in terms of bidirectional similarities for all of the videos. A value of $1.870$ for video $4$ shows the retargeted video is very similar to the source.

Next, we compare the deep features between the source and retargeted stereo video frames. For this purpose, we compute the difference between the VGG19 \cite{simonyan2014very} features extracted from the source and target frames, respectively. Table \ref{table2} depicts the quantitative comparison of the VGG19 features. Video \#1 obtained the best results of $0.3262$, indicating that the retargeted stereo video is similar to its source.

{To evaluate depth preservation, we compute the depth distortion with the DDr metric for the video shown in Figure \ref{d4}. The DDr results are reported in the figure for each aspect ratio, together with the illustration of the corresponding disparity maps. It can be observed that the distortion is lower for smaller size reduction (size reduction of $30\%$, $20\%$), as compared to more extreme cases ($50\%$ and $150\%$). For horizontal size reduction of 20\%, a low distortion ratio of $DDr=0.112$  is reported.}

\begin{figure}[b!]
\vspace{-1em}
\hspace{0.2cm} Input Frames \hspace{0.7cm} LS \hspace{0.4cm} w/o CoSD \hspace{0.01cm} w/o Trans \hspace{0.1cm} ours 

\includegraphics[width=1\linewidth]{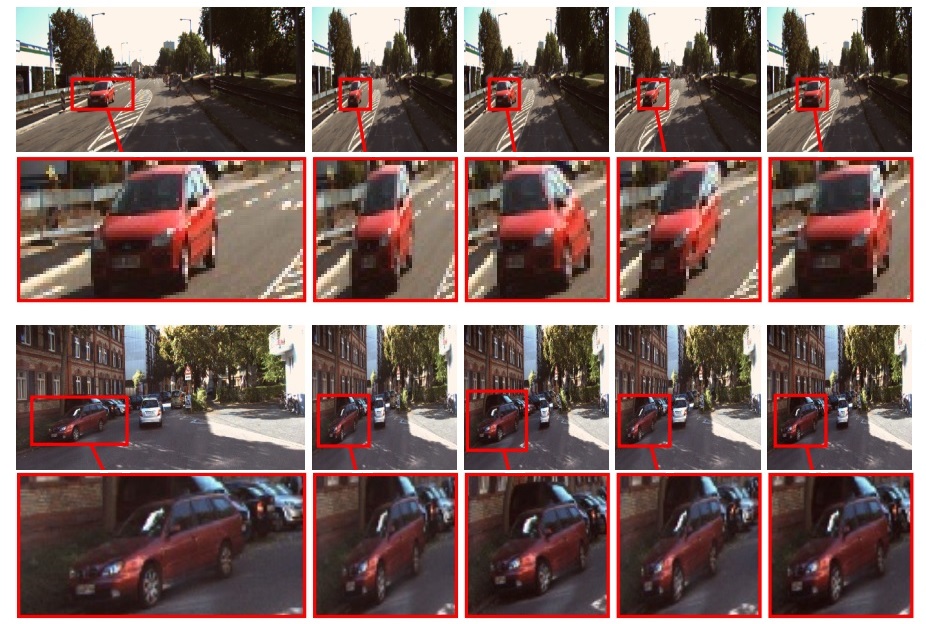}
  \vspace{-1.5em} 
  \caption{{Ablation study. Performance comparison of our model without using CoSD (\textit{w/o CoSD}), without using the SVT (\textit{w/o Trans}), and with all modules (\textit{ours}). Videos are selected from the KITTI stereo 2015 \cite{menze2015joint} dataset.}}
\label{d6}
\end{figure}

\begin{table}
		\centering
			\caption{{Ablation study. Comparison of the similarity between the input and retargeted videos based on the VGG19 \cite{simonyan2014very} features. The results are for 3 cases: without
(\textit{w/o CoSD}) CoSD saliency detection, without (\textit{w/o Trans}) Transformer block, and will all of the blocks (\textit{with all}). The best results are shown in \textbf{bold}.}}
            \vspace{-0.5em}
		\scalebox{0.84}{
		\begin{tabular}{l@{\hspace{25pt}} *{14}{c}}
			\hline 
            {\textbf{Video No.}} & {\textbf{\#1}} & {\textbf{\#2}} & {\textbf{\#3}}& {\textbf{\#4}} & \textbf{Avg} \\
			\hline


			w/o CoSD  & 0.8755 & 0.9400 & 1.3301 & 0.8497 & 0.9988\\
			w/o Trans  & 0.7737 & 0.9221 & 1.2107 & 0.7087 & 0.9038\\
			with all  & \textbf{0.5584} & 0\textbf{0.5971} & \textbf{0.6244} & \textbf{0.3515} & \textbf{0.5328}\\
			\hline 
		\end{tabular}}
            \vspace{-1.5em}
	\label{tableAb}
\end{table}

\subsection{Ablation Study}
The key idea of the proposed framwork is to preserve the salient regions during the retargeting process. Without detecting the salient parts, our method works like linear scaling. Therefore, instead of removing the whole saliency detection block that combines CoSD, object detection, and disparity information for the ablation study, we only remove the CoSD module to investigate how it affects the retargeting process. We further ablate with removing the SVT from our model and seeing its impact. Figure \ref{d6} shows the results of the ablation study. With the CoSD module removed, the results show that the salient parts are not well detected, affecting the final retargeting results. The deformation of the main object is apparent in both examples. Without the CoSD module, the other parts of the frame are not affected much, but the main objects are deformed. The situation differs when {SVT} is removed. The training process is affected without the Transformer, and all parts of the frames are affected. More detailed ablation studies are reported in the supplementary materials. {In Table \ref{tableAb}, we study the influence of CoSD and SVT modules for stereo video retargeting based on the VGG19 feature comparison. As expected, the results show that without the CoSD module (\textit{w/o CoSD}) or SVT module (\textit{w/o Trans}), the results degrade significantly, with the feature difference score increased to almost double in some cases.} {In addition, in the supplementary materials, we show that our method's performance slightly degrades without uthe use of the disparity information.} {From Table \ref{tableAb} and Figure \ref{d6}, we can conclude that removing the CoSD model is more detrimental than removing the SVT model. The performance of the other blocks will be affected without accurately detecting the salient regions. In the future, we will explore assigning more weights to the attention generation in the Transformer block, when fusing SVT and CoSD so that our approach would depend less on saliency detection.}

\begin{figure}[tb] 
\centering
\includegraphics[width=1.0\linewidth]{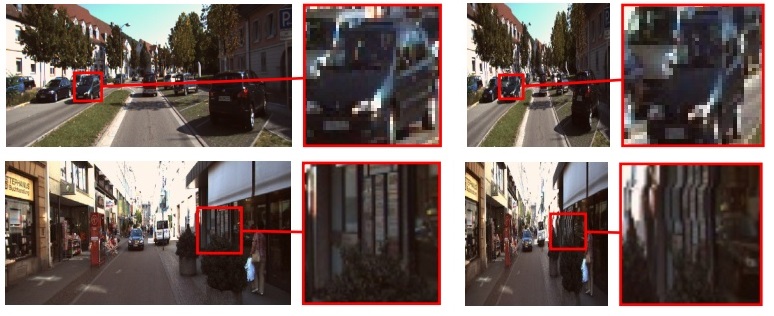}
  \vspace{-2em}
  \caption{{Example of failure cases due to: (top) videos with many salient objects and (bottom) wrong detection of salient objects.}}
\label{fail}
\vspace{-1em}
\end{figure}




\section{Conclusions}
In this paper, we proposed a new unsupervised stereo video retargeting method. Our model detects the salient objects, and shifts and warps all the objects in a manner that gives more attention to the salient parts of the stereo frames. We use 1D convolution for shifting the salient objects and design a stereo video Transformer (SVT) to assist the retargeting process. In addition, we reconstruct the source frames from retargeted ones using the PAM module, and a convolutional reconstruction block is used to train the model in an unsupervised manner. Extensive quantitative and qualitative experimental results on the KITTI stereo 2012 and 2015 datasets demonstrate the effectiveness of our proposed stereo video retargeting framework in preserving spatial, temporal, and disparity information. 

However, our method fails in some extreme retargeting cases ($>50\%$ reduction in size). For example, when there are significant salient objects or complex scenes, the shape of some salient objects can be deformed. The first row of Figure \ref{fail} shows an example of this case. The black car shape is deformed due to the existence of other salient objects. The other case is related to the failure of the saliency detection method. The two trees in this scene are mistakenly detected as salient objects. 
To overcome these limitations, additional constraints might be required to preserve the shapes for extreme retargeting cases, which warrants further investigation. 
Another aspect of video retargeting worth investigating is formulating a benchmark metric for evaluating the performance of retargeted stereo videos.

\noindent \textbf{Acknowledgements.} This work is supported by the Scientific and Technological Research Council of Turkey (TUBITAK) 2232 Leading Researchers Program, Project No. 118C301.
 
{\small
\bibliographystyle{ieee_fullname}
\bibliography{egbib}
}

\end{document}